\documentclass[conference,10pt]{IEEEtran}
\usepackage{cite}
\usepackage{amsmath,amssymb,amsfonts}
\usepackage{algorithmic}
\usepackage{graphicx}
\usepackage{textcomp}
\usepackage{xcolor}
\usepackage{refstyle}
\usepackage[numbers]{natbib}
\def\BibTeX{{\rm B\kern-.05em{\sc i\kern-.025em b}\kern-.08em
    T\kern-.1667em\lower.7ex\hbox{E}\kern-.125em}}
\usepackage[colorlinks=true,linkcolor=blue]{hyperref}
\usepackage{caption}
\usepackage{subcaption}
\usepackage{booktabs}
\usepackage{siunitx}

\newcommand{\bth}{\boldsymbol\theta}

\newcommand{\bw}{\mathbf{w}}

\begin{document}
\hypersetup{draft}
\bstctlcite{IEEEexample:BSTcontrol}

\title{HYPPO: A Surrogate-Based Multi-Level Parallelism Tool for Hyperparameter Optimization}

\author{
    \IEEEauthorblockN{
        Vincent Dumont\IEEEauthorrefmark{1}\IEEEauthorrefmark{8},
        Casey Garner\IEEEauthorrefmark{2}\IEEEauthorrefmark{1},
        Anuradha Trivedi\IEEEauthorrefmark{3}\IEEEauthorrefmark{1},
        Chelsea Jones\IEEEauthorrefmark{4}\IEEEauthorrefmark{1},
        Vidya Ganapati\IEEEauthorrefmark{5}\IEEEauthorrefmark{1},\\
        Juliane Mueller\IEEEauthorrefmark{1},
        Talita Perciano\IEEEauthorrefmark{1},
        Mariam Kiran\IEEEauthorrefmark{1},
        and Marc Day\IEEEauthorrefmark{6}
    }\\
    \IEEEauthorblockA{
        \IEEEauthorrefmark{1}
            \textit{
                Computing Sciences Area,
                Lawrence Berkeley National Laboratory,
                Berkeley, CA 94720, USA
            }\\
        \IEEEauthorrefmark{2}
            \textit{
                Department of Mathematics,
                University of Minnesota Twin Cities,
                Twin Cities, MN 55414, USA
            }\\
        \IEEEauthorrefmark{3}
            \textit{
                Department of Biomedical Engineering,
                Georgia Institute of Technology and Emory University,
                Atlanta, GA 30332, USA
            }\\
        \IEEEauthorrefmark{4}
            \textit{
                Department of Statistical Sciences \& Operations Research,
                Virginia Commonwealth University,
                Richmond, VA 23284, USA
            }\\
        \IEEEauthorrefmark{5}
            \textit{
                Department of Engineering,
                Swarthmore College,
                Swarthmore, PA 19081, USA
            }\\
        \IEEEauthorrefmark{6}
            \textit{
                Computational Science Center,
                National Renewable Energy Laboratory,
                Golden, CO 80401, USA
            }\\
    }\\
    
    \IEEEauthorblockA{
        \IEEEauthorrefmark{8}
            Corresponding Author:  \href{mailto:vincentdumont11@gmail.com}{vincentdumont11@gmail.com}
    }
    
}
\maketitle
\begin{abstract}
We present a new software, HYPPO, that enables the automatic tuning of hyperparameters of various deep learning (DL) models. Unlike other hyperparameter optimization (HPO) methods, HYPPO uses adaptive surrogate models and directly accounts for uncertainty in model predictions to find accurate and reliable models that make robust predictions. Using asynchronous nested parallelism, we are able to significantly alleviate the computational burden of training complex architectures and quantifying the uncertainty. HYPPO is implemented in Python and can be used with both TensorFlow and PyTorch libraries. We demonstrate various software features on time-series prediction and image classification problems as well as a scientific application in computed tomography image reconstruction. Finally, we show that (1) we can reduce by an order of magnitude the number of evaluations necessary to find the most optimal region in the hyperparameter space and (2) we can reduce by two orders of magnitude the throughput for such HPO process to complete.
\end{abstract}

% Introducing the problem and our proposed solution
\section{Introduction}
% \mgt{We should keep in mind that this paper is for supercomputer audience and the following questions should be answered in the introduction:
% \begin{enumerate}
%     \item Whose problem are you solving?
%     \item What are the benefits?
%     \item What has been done to solve that before?
%     \item Why couldn't they solve it?
%     \item What is my solution?
%     \item How do we validate it?
%     \item Show the benefit again?
% \end{enumerate}
% }

%\cite{Muller2020}
Deep learning models are increasingly used throughout the sciences to achieve various goals, including the acceleration of time-consuming computer simulations, for making predictions in cases where the underlying physical behavior are poorly understood, for deriving new insights into physical relationships, and for decision support. E.g., in~\cite{pathak2020using}, a low-fidelity simulation model is augmented with predictions made by a U-Net to obtain high-fidelity data at a fraction of the computational cost of running a high-fidelity simulation. 
In~\cite{Muller2020}, DL models are used to make long-term predictions of groundwater levels in California in order to enable computationally fast predictions of future water availability. 
\cite{10.1145/3437359.3465597} used convolutional neural networks (CNNs) for anomaly detection in scientific workflows. 
DL models can also be used to co-optimize hardware acquisition parameters with image reconstruction in computational imaging, as in~\cite{cheng_illumination_2019, robey_optimal_2018}. \cite{LAPOINTE2020133} used neural networks to predict the CPU times of Ordinary Differential Equation (ODE) solvers, thus enabling the optimal selection of ODE solvers.

One obstacle to using DL models in new scientific applications is the decision about  which hyperparameters (e.g., number of layers, nodes per layer, batch size) should be used. 
The hyperparameters impact the accuracy of the DL model's predictions and should therefore be tuned. 
However, the predictive performance is subject to variability that arises due to the use of stochastic optimizers such as stochastic gradient descent (SGD) for training the models. 
This variability must be taken into account during hyperparameter optimization (HPO) in order to achieve models that produce accurate predictions reliably. 

There are many libraries defined for hyperparameter tuning. Each of them has different strengths, compatibility, and weakness. 
Some of the frequently used libraries are Ray-Tune which supports state-of-the-art algorithms such as population-based training~\cite{jaderberg2017population}, Bayes Optimization Search (BayesOptSearch)~\cite{bayesopt, 10.5555/2999325.2999464}; Optuna, which provides easy parallelization~\cite{akiba2019optuna}; Hyperopt, which works both serial and parallel ways and supports discrete, continuous, conditional dimensions~\cite{10.5555/3042817.3042832}; Polyaxon, which is for large scale applications~\cite{parameterlib}.
 
However, little work discusses how the prediction uncertainty of deep learning models can be reduced with these approaches. This paper describes a new software tool, HYPPO, for automated HPO that is based on ideas from surrogate modeling, uncertainty quantification, and bilevel black-box optimization. In contrast to other HPO methods such as DeepHyper~\cite{8638041} and MENNDL~\cite{10.1145/2834892.2834896}, our HPO implementation directly takes the prediction variability into account to deliver \textit{highly accurate and reliable} DL models. 
We demonstrate the approach on the CIFAR10 image classification problem, time series prediction and a computed tomography image reconstruction application.
We take full advantage of high-performance computing (HPC) environments by using asynchronous nested parallelization techniques to efficiently sample the high-dimensional hyperparameter space.

% In this work, we include a measure of prediction uncertainty in the optimization cost function, and we use surrogate modeling to identify hyperparameters that yield accurate and reliable predictions. 

\noindent
\textbf{Relevance, impact, and contribution to the literature:
\begin{enumerate}
    \item Automated and adaptive HPO for DL models that directly accounts for uncertainty in model predictions;
    \item Significant decrease in the number of evaluations necessary to identify the most optimal region in the hyperameter space;
    \item Nested parallelism significantly accelerates time-to solution (up to two orders of magnitude);
    \item Compatible with both TensorFlow and PyTorch;
    \item HYPPO can run on large compute clusters as well as laptops and is able to exploit GPU and CPU;
    \item Provides reliable and robust models.
\end{enumerate}
}

\section{Motivation}
Much work on hyperparameter search extensively studies applications in CNNs for image recognition \cite{xiao2020efficient} or general deep learning models \cite{8638041}. 
Further development of various hyperparameter search libraries such as RLlib \cite{pmlr-v80-liang18b}, Hyperband \cite{10.5555/3122009.3242042}, DeepHyper \cite{8638041}, and even hand-tuning help iterate through multiple settings repeatedly until finding fairly well-performing hyperparameters.
However, deep learning models can lead to multiple predictions with varying range of values, even after tuning the hyperparameters.
Current hyperparameter tuning libraries often ignore this challenge of prediction variability.
In this work we particularly target this problem of uncertainty quantification, by building in uncertainty when tuning hyperparameters. The ability to quantify the prediction uncertainty of DL models has the advantage that we can obtain confidence bounds for  the DL model's  predictions, as illustrated in two examples in Figure~\ref{time_series_var}. The additional information about prediction uncertainty can be very advantageous in various applications. For instance, instead of fully trusting a model architecture because of good predictive  performance (which can be interpreted as a  single realization of a stochastic process), UQ provides additional information on how reliably that  architecture performs. If, for example, one was to predict future groundwater levels to enable sustainable groundwater management  (see~\cite{Muller2020}), having an idea about the prediction variability  enables managers to develop strategies that are based on best, worst, and average case performance information and thus lead to robust management decisions. 
Similarly, in classification tasks, scientists may not only be interested in the class with highest expected probability, but also in information such as the second or third most probable class membership as well as uncertainties of those memberships which will allow  further insights into the significance of differences between class memberships. In particular, in scientific applications where determining the correct class may have huge negative impacts if done incorrectly, this additional information is highly relevant. 
The main contributions of HYPPO include (1) an automated and adaptive search for hyperparameters using uncertainty quantification to evolve the best deep neural network solutions for both PyTorch and Tensorflow codes, and (2) demonstrate the scalability of our solution using high-performance computing (HPC) to leverage multiple threads and parallel processing to improve the overall search time for hyperparameters.

Our experiments cover time series prediction, image classification, and image reconstruction, and present how surrogate modeling can help to find optimal deep learning architectures that minimize the uncertainty in the solutions.

% Explain background idea of using surrogate modeling
\section{Bi-Level Optimization Problem}
Mathematically, we formulate the HPO problem as a bilevel bi-objective optimization problem:
\begin{eqnarray}
&\displaystyle  \min_{\bth, \bw^*} & (\ell_1(\bth, \bw^*;\mathcal{D}_\text{val}),\ell_2(\bth, \bw^*;\mathcal{D}_\text{val})) \label{eq:up}\\
&\displaystyle \text{s.t. } &\bth\in\Omega \label{eq:theta}\\
&&\displaystyle  \bw^*\in\arg\min_{\bw\in\mathcal{W}}L(\bw; \bth, \mathcal{D}_\text{train}).\label{eq:low}
\end{eqnarray}
Here, $\bth$ represents the set of hyperparameters to be tuned, $\Omega$ describes the space over which the parameters are tuned (in our case an integer lattice), and $\ell_1$ and $\ell_2$  represent the loss and loss variability of  the trained model, respectively. 
Ideally, we want to find optimal $\bth$ such that both $\ell_1$ and $\ell_2$ are minimized simultaneously.
Note that most commonly only a single objective that represents a mean squared loss is minimized and the variability is not accounted for. 

The values of $\ell_i, i = 1,2$, for each hyperparameter set for the validation dataset $\mathcal{D}_\text{val}$ can be computed only after solving the lower level problem (Eq. \ref{eq:low}), in which the weights and biases $\bw^*$ are obtained by training (e.g., with  SGD). Solving this lower level problem is, depending on the size of the training dataset $\mathcal{D}_\text{train}$ and the size of the architecture, computationally expensive. The loss variability arises when the same architecture is trained multiple times because different solutions $\bw^*$ are obtained (see Section~\ref{sec:sw}, Feature 1, for details), and thus different values for the loss $\ell_1$ at the upper-level result. 
Our goal is to find optimal hyperparameters $\boldsymbol{\theta}^*$ that yield high predictive accuracy (low $\ell_1$ values) and low prediction variability (low $\ell_2$ values).  

In order to tackle this problem, we interpret the HPO problem as a computationally expensive black-box problem in which the black-box evaluation corresponds to a stochastic evaluation of $\ell_1$ with variability $\ell_2$. 
Furthermore, to alleviate the complexities of solving a bi-objective optimization problem, we use a single-objective reformulation in which a weighted sum of $\ell_1$ and $\ell_2$ is minimized and the weights corresponding to each reflect their respective importance. 

To solve our HPO problem, we use ideas from the surrogate model-based optimization algorithm presented in~\cite{Muller2020} and extend it to consider the model uncertainty. 
We make full use of HPC by exploiting asynchronous nested parallelism strategies which significantly reduces the time to finding optimal architectures.

\subsection{Surrogate Model-based HPO}

Surrogate models such as radial basis functions (RBFs)~\cite{Powell1992} and Gaussian process models (GPs)~\cite{Matheron1963} have been widely used in the derivative-free optimization literature. These models are used to map the parameters $\bth$ to the model performance metrics, $\ell_i(\bth)$ (here the dependence of $\ell_i$ on $\bw^*$ is implied by means of interpreting the lower level problem as black-box evaluation). The surrogate models are computationally cheap to build and evaluate and thus enable an efficient and effective exploration of the search space~\cite{Jones2001}. Adaptive optimization algorithms using global surrogate models usually consist of the following three steps: 
\begin{enumerate}
    \item Create an initial experimental design and evaluate the costly objective function to obtain input-output data pairs. 
    \item Build a surrogate model based on all data pairs.
    \item Solve a computationally cheap auxiliary optimization problem on the surrogate model to select new point(s) at which the expensive function is evaluated and go to step 2. 
\end{enumerate}
In order to apply these methods to HPO, special sampling strategies that respect the integer constraints of the hyperparameters are needed (see~\cite{Muller2020} for details). 
Moreover, the iterative sampling strategy must take into account the performance variability, which can be done by defining suitable model performance metrics. 
%[unsure if itll stay: Moreover, optimization is subject to the curse of dimensionality and new sensitivity analysis methods that work on integer-constrained parameters can help us to understand the importance and interactions of hyperparameters, thus  potentially decreasing the  search space.]

%In the following subsections, we provide details on (1) how we address the problem of prediction variability,  and (2) how asynchronous nested parallelism enables us to significantly accelerate HPO.

% Describe software's design and implemented features
\section{HYPPO Software: Design \& Features}\label{sec:sw}
In this section, we provide details of specific features of our HYPPO software and implementation. We describe our approach to accounting for the prediction variability during HPO, the proposed surrogate modeling and sampling approaches, and the asynchronous nested parallelism that enables us to speed up the calculations significantly.

%\subsection*{Feature 1: Integer-Based Low-Discrepancy Sampling}
%\input{chelsea}
\subsection*{Feature 1: Uncertainty Quantification} 
%\mgt{CASEY: Describe approach to evaluate uncertainty from hyperparameter sets and propagate uncertainty throughout the training and during surrogate modeling.}
%\mgt{(Start of Casey's 2nd-ish draft)}

%In this section, we expound on how uncertainty quantification (UQ) is employed and applied in conjunction with surrogate model based HPO. 
HYPPO employs and applies uncertainty quantification (UQ) in conjunction with surrogate model-based HPO.
The prediction variability present in Eq.~(\ref{eq:low}) is epistemic where the variability observed manifests itself because Eq.~(\ref{eq:low}) is a non-convex problem being solved in-exactly by stochastic algorithms. 
Non-convexity limits knowledge of global optimality, while in-exactly solving the lower level problem gives no guarantee that the $\bw^*$  used to evaluate the loss $\ell_1$ in Eq.~(\ref{eq:up}) is a stationary point to Eq.~(\ref{eq:low}). 

The body of potential UQ approaches to bound the performance variability for a fixed hyperparameter set $\boldsymbol{\theta}$ is thinned significantly by the nature of the problem. Since evaluating the loss is an expensive black-box function, any UQ method that depends on heavy sampling  quickly becomes computationally intractable.  Furthermore, distributional information for the performance value  based on $\bw^*$ is unknowable beyond trivial cases because such knowledge is tied to the local optima of Eq.~(\ref{eq:low}), which in general are neither computed during training nor enumerated.
An additional blockage to distributional knowledge comes from the possibility that different global optima of Eq.~(\ref{eq:low}) likely yield different loss function values in Eq.~(\ref{eq:up}).
Finally, flexible UQ approaches capable of handling various DL models are essential for the broad applicability of our software. 
Considering the inherent difficulties and desired properties delineated, the central framework we utilize for UQ is Monte Carlo (MC) dropout \cite{Gal2016}.

MC dropout was developed in the form presented by Gal and Ghahramani \cite{Gal2016,Gal_PhD} as an alternative derivation of \textit{model averaging} from \cite{Srivastava2014}. 
The key idea behind MC dropout is to use standard dropout during testing. 
Dropout is a technique often applied to prevent a network from overfitting and thus improve its generalizability~\cite{Srivastava2014, DO_survey}. This is achieved by taking each node out of the network with some probability $p$ during training and then multiplying the network's weights by $p$ during test time. 
Equivalently, as done in PyTorch and TensorFlow dropout layers, nodes can be dropped with probability $p$ and the value of the remaining nodes scaled by $1/(1-p)$ during training resulting in no scaling of the weights during testing. 
So, MC dropout applies dropout on a network at test time, i.e., when the input is evaluated by forward-propagation, each node is dropped out with probability $p$ and the output of the remaining nodes is scaled by $1/(1-p)$. 
Therefore, forward-propagating the same input through a network using dropout will produce different outputs on each pass from which we can obtain a measure of variability for the output. Formulating this mathematically, let $\mathbf{y}_t(\mathbf{x})$ be the output of a DL model for input $\mathbf{x}$ on the $t$-th pass through the network using MC dropout. 
From $T$ total passes through the network, we can compute a sample mean for the output of input $\mathbf{x}$ as,
\begin{equation}\label{eqn:DO_mean}
\mathbf{\bar{y}}(\mathbf{x}) = \frac{1}{T}\sum_{t=1}^{T} \mathbf{y}_t(\mathbf{x}),
\end{equation}
and a sample variance,
\begin{equation}\label{eqn:DO_var}
\sigma_{\text{model}}^2 = \frac{1}{T} \sum_{t=1}^{T} \left( \mathbf{y}_t(\mathbf{x}) - \bar{\mathbf{y}}(\mathbf{x})\right)^2,
\end{equation} 
where the squared terms in Eq.~(\ref{eqn:DO_var}) are squared element-wise, i.e. $\left[(\mathbf{a} - \mathbf{b})^2\right]_j = ({a}_j - {b}_j)^2$ for all $j$ where $a_j$ is the $j$-th element of $\mathbf{a}$.  
%$||\cdot||$ is the standard Euclidean norm and is equivalent to the absolute value of its argument if the output of the model is one-dimensional. 
A crucial observation is that MC dropout derives variability measures from repeated evaluations of a DL model rather than large numbers of repeated training of the same model or through the construction of another network. 
Thus, it satisfies the general restrictions imposed by the characteristics of our loss functions in Eq.~(\ref{eq:up}).

%We note that MC dropout requires no additional training nor alteration to the DL model and is of little computational expense compared to training the same architecture tens of times in order to obtain a sample estimate of the prediction variability. Thus, it satisfies the general restrictions posed by our objective function (\ref{eq:up}). 
\begin{figure*}[t]
\centering
\begin{subfigure}{.5\textwidth}
  \centering
  \includegraphics[width=\columnwidth]{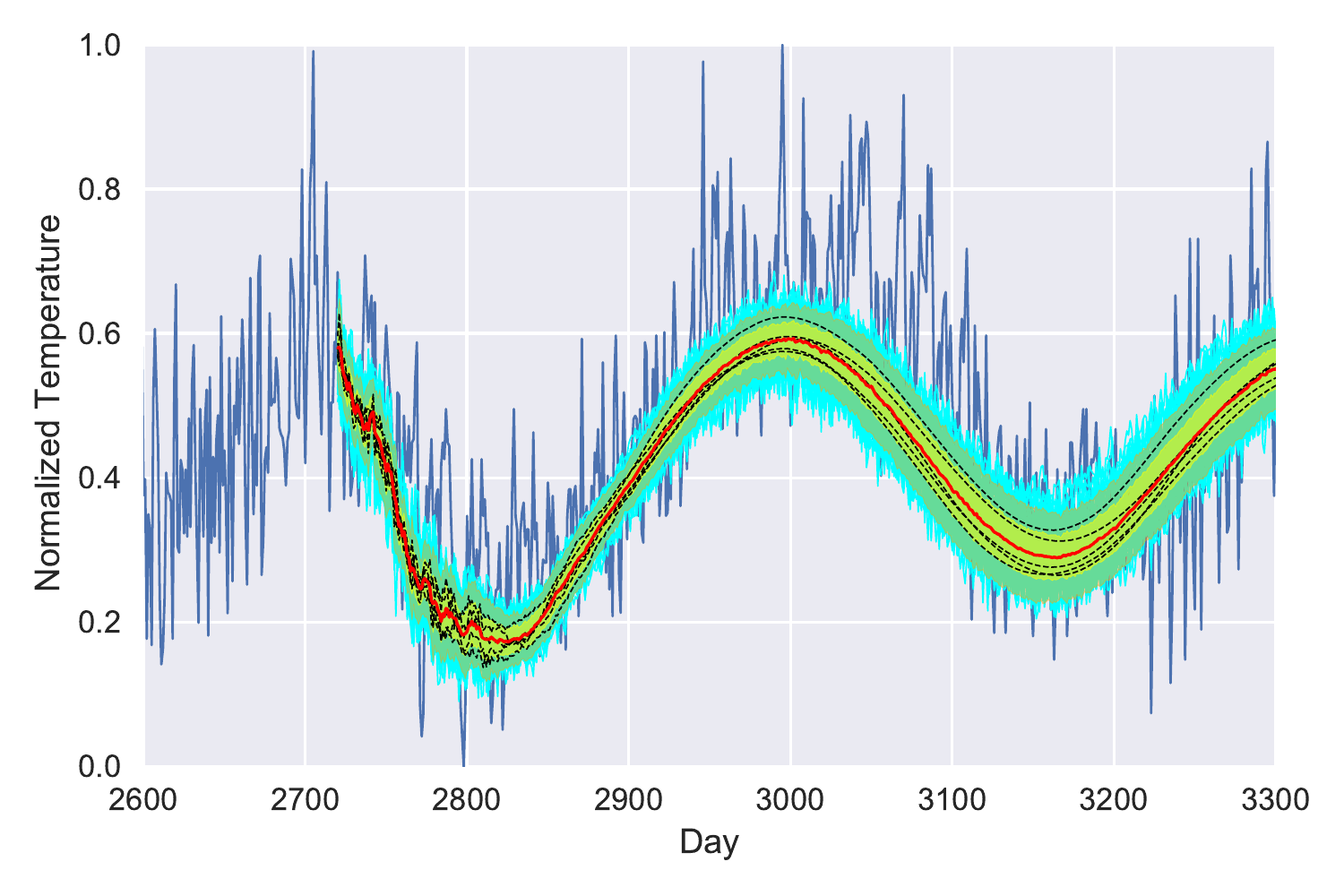}
  \caption{Time series prediction of temperature in Melbourne, Australia}
  \label{fig:prediction}
\end{subfigure}%
\begin{subfigure}{.5\textwidth}
  \centering
  \includegraphics[width=\columnwidth]{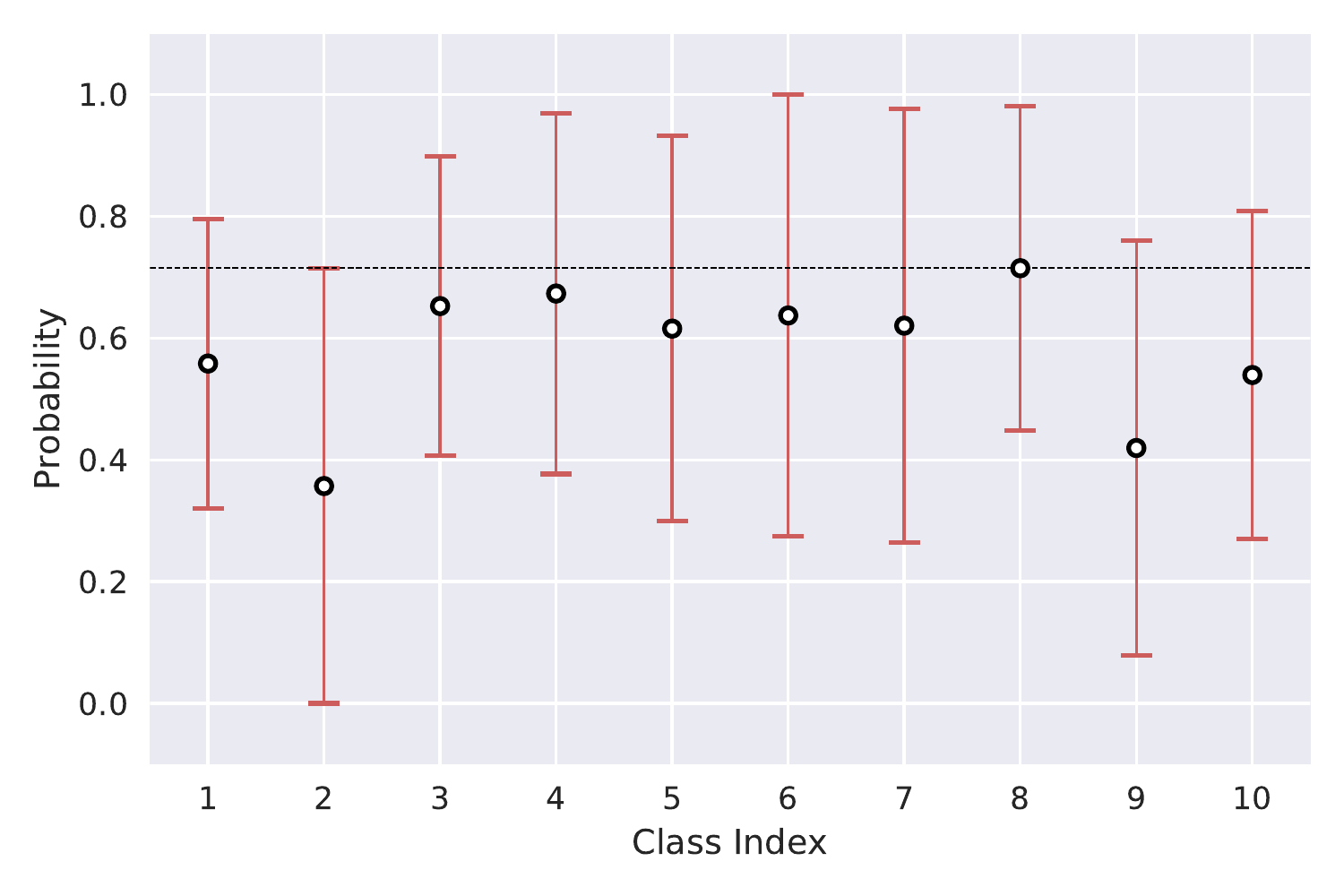}
  \caption{CIFAR10 image classification}
  \label{fig:classification}
\end{subfigure}
\caption{Application of Monte Carlo dropout to a DL models. In Fig. \ref{fig:prediction}, the yellow and green regions represent the $\pm 1$ and $2$ standard deviations from the mean (red line). The cyan lines are the MC dropout predictions and the black dashed lines are the mean predictions from each of the 5 independent trial models. 
  The DL model used ``prediction-on-prediction'', i.e., the model's predicted values were utilized in making future predictions. In Fig. \ref{fig:classification}, the confidence interval around the mean probability for each class is shown for a single input image. While the right class (number 8) is correctly identified, the uncertainty of the output probability provides important information with regard to the stability and accuracy of the trained model.}
  \label{time_series_var}
\end{figure*}

Our UQ approach applies MC dropout to a relatively small set of trained models with  the same architecture to form a weighted average. 
Combining MC dropout with repeated training of the same architecture balances the explanatory power of repetitive training with the computational frugality of MC dropout.
Assume we have $N$ trained models with the same architecture $\boldsymbol{\theta}$ for which we generate $T$ dropout masks each. 
We compute the expected output of a trained model with the same architecture as,
\begin{equation}\label{eqn:unc_mean}
\boldsymbol{\mu}_{\text{pred}}(\mathbf{x}) = \frac{w_T}{N}\sum_{i=1}^{N}\boldsymbol{y}^{i}(\mathbf{x}) + \frac{w_D}{NT}\sum_{j=1}^{N}\sum_{t=1}^{T} \boldsymbol{y}_t^j(\mathbf{x}),
\end{equation}
where $\boldsymbol{y}^i(\mathbf{x})$ is the output of the trained model $i\in\{1,\hdots,N\}$ for input $\mathbf{x}$ without dropout at test time, $\boldsymbol{y}_t^j(\mathbf{x})$ is the output of the dropout model $t\in \{1,\hdots,T\}$ for trained model $j\in\{1,\hdots,N\}$, and $w_T \geq 0$, $w_D >0$ with $w_T + w_D  = 1$. 
The variance of the expected output is computed as,
\begin{multline}\label{eqn:model_var}
\boldsymbol{V}_{\text{model}}(\mathbf{x}) = \frac{w_T}{N}\sum_{i=1}^{N}\left( \boldsymbol{\mu}_{\text{pred}}(\mathbf{x}) - \boldsymbol{y}^i(\mathbf{x})\right)^2\\  + \frac{w_D}{NT}\sum_{j=1}^{N}\sum_{t=1}^{T} \left( \boldsymbol{\mu}_{\text{pred}}(\mathbf{x}) - \boldsymbol{y}_t^j(\mathbf{x})\right)^2.
\end{multline}
Note that Eqs.~(\ref{eqn:unc_mean}) and~(\ref{eqn:model_var}) are weighted averages of the computed mean and variances of the trained models and the MC dropout models. 
These equations enable micro-predictions of the performance uncertainty of a model for a given input. 
The weighted average parameters $w_T$ and $w_D$, as well as the number of iterations $T$ of MC dropout per trained model, are user-defined settings.
An example of our modified MC dropout approach can be seen in Figure~\ref{time_series_var} for time series prediction of daily temperature in Melbourne, Australia, and CIFAR10 image classification.
Here $w_T = w_D = 0.5$ and $T = 30$, which correspond to our default values. The robustness of the model predictions can be measured, for example, in the average  width of the uncertainty bands associated with each day's temperature prediction and with each class membership probability, respectively. 

% \begin{figure}[t]
% \begin{center}
% \includegraphics[width=\linewidth]{imgs/Time_Series_Temp.jpg}
% \caption{\textcolor{red}{move to introduction - use model with better hyperparameters}Application of Monte Carlo dropout to a DL model for time-series prediction. The red lines are the MC dropout predictions, the green line is the mean prediction,  %$\boldsymbol{\mu}_{\text{pred}}$, 
% and the yellow and purple lines give $\pm 1$ and $2$ standard deviations from the mean. %, $\sqrt{\boldsymbol{V}_{\text{model}}}$, from $\boldsymbol{\mu}_{\text{pred}}$. 
% The DL model used ``prediction-on-prediction'', i.e., the model's predicted values were utilized in making future predictions. }
% \label{time_series_var}
% \end{center}
% \end{figure}

 Eqs.~(\ref{eqn:unc_mean}) and~(\ref{eqn:model_var}) can then be used to obtain confidence intervals for the loss function value corresponding to a given hyperparameter set $\boldsymbol{\theta}$.
  For example, if $ \ell_1\left( \boldsymbol{\theta}; \bw^*, D_{\text{val}}\right)$ is the mean-squared loss function with validation data set $\mathcal{D}_{\text{val}} = \left\{ (\mathbf{x}^i, \mathbf{z}^i)\right\}_{d=1}^{D}$, then we approximate the expectation  over $\bw^*$  by 
\[
 \mathbb{E}_{\bw^*}\left(\ell_1\right) \approx \frac{1}{2D}\sum_{d=1}^{D}\big|\big| \mathbf{z}^i - \boldsymbol{\mu}_{\text{pred}}(\mathbf{x}^i) \big|\big|^2.
 \] 
% by plugging in $\boldsymbol{\mu}_{\text{pred}}$ into the outer loss function. 

The intuition is that $\boldsymbol{\mu}_{\text{pred}}$ serves as a good estimate of the actual expectation of the output over all possible trained models with architecture $\boldsymbol{\theta}$. 
Now, from computing the outputs of each of the trained models, $\mathbf{y}^i(\mathbf{x}^d)$, and their many dropout masks, $\mathbf{y}^i_t(\mathbf{x}^d)$, for all of the inputs $\mathbf{x}^d\in \mathcal{D}_{\text{val}}$, we are able to compute the value of $\ell_1$ for each of the models utilized to compute $\boldsymbol{\mu}_{\text{pred}}$. Taking the standard deviation of these $N + TN$ outer loss values gives an estimate for the variability $\ell_2$ in the outer loss function for a model with architecture $\boldsymbol{\theta}$. Letting the value of $\ell_1$ computed from $\boldsymbol{\mu}_{\text{pred}}$ be the center and using the aforementioned standard deviation as a radius, we obtain a confidence interval for the outer objective. %In order to compute confidence intervals with the HYPPO software, \textit{uq\textunderscore on} must be set to \textit{True} in the \textit{uq}-section of the configuration file. 

Upon obtaining confidence intervals and output variances using Eq.~(\ref{eqn:model_var}) for $\ell_1$, we desire to utilize this additional uncertainty information in the HPO. 
Our software uses uncertainty in different ways.
% This can be achieved through the software in various manners. 
The first option for integrating uncertainty is to use the surrogate models with the value of $\ell_1$ taken as its output when computed with $\boldsymbol{\mu}_{\text{pred}}$. 
The RBF and GP based optimization approaches available in the software (see next section for details) would then run as expected with confidence intervals being computed and tabulated for all new hyperparameter settings. %\textcolor{blue}{Juli says: move to software section: Assigning {\it uq \textunderscore on: True} and {\it uq\textunderscore hpo: False} \textcolor{blue}{(to be changed to {\it ensemble\textunderscore on: False})} in the configuration file implements this approach with either surrogate model. }

The second  option is to use the confidence intervals to construct an ensemble of RBFs to perform the surrogate modeling; the {\it RBF ensemble approach} uses multiple RBFs % of the form,
%\begin{equation}\label{eqn:rbf}
%  m_{\text{RBF}}(\boldsymbol{\theta}) = \sum_{j=1}^{n} \lambda_j \phi\left(|| \boldsymbol{\theta} - \boldsymbol{\theta}_j ||_2\right) + p(\boldsymbol{\theta}),
%\end{equation}
%\noindent where $\boldsymbol{\theta}_j$ are the sets of hyperparameters for which $\ell$ has previously been evaluated, $\phi(r) = r^3$, and $p(\boldsymbol{\theta}) = \beta_0 + \boldsymbol{\beta}^T \boldsymbol{\theta}$. A system of equations (see (6) in \cite{Muller2020}), dependent on the value of $\ell$ at the $\boldsymbol{\theta}_j$'s, is solved to compute $\lambda$, $\beta_0$ and $\boldsymbol{\beta}$. 
that are generated from the confidence intervals by selecting uniformly at random from the extremes of these intervals, i.e., the lower bound, center, and upper bound. %The values sampled from the confidence intervals are used to update the right-hand side of (6) in \cite{Muller2020} thus generating multiple response surfaces.
Each response surface provides an estimate for the value of $\ell_1$ at potential hyperparameter candidates, $\boldsymbol{\theta}^C_i$ (see Feature 2 for description of their generation), and we compute a mean $\mu(\boldsymbol{\theta}^C_i)$ and standard deviation $\sigma(\boldsymbol{\theta}^C_i)$ over all ensemble predictions for $\ell_1$ at each candidate point.
The selection of the best candidate point (and thus new hyperparameters to be evaluated) is determined by which candidate point minimizes a weighted sum of its distance to previously evaluated points and,
\begin{equation}\label{eq:weight_avg_in_rbfe}
\mu(\boldsymbol{\theta}^C_i) + \alpha \sigma(\boldsymbol{\theta}^C_i),
\end{equation}
where $\alpha \in [-2,2]$. Note that if $\alpha = 0$, then only the mean value of $\ell_1$ predicted by the ensemble is considered when determining the next hyperparameter setting. 
If $\alpha = 2$, a ``pessimistic'' approach is taken by penalizing candidate points with large prediction variability. 
On the other hand, by setting $\alpha = -2$, an ``optimistic'' stance is taken by preferring  candidate points with relatively good mean performance and large standard deviation that may lead to significant improvements. %\textcolor{blue}{Juli says: maybe put this into software section: To implement the {\it RBF ensemble approach} during the surrogate modeling, {\it uq\textunderscore on} and {\it uq\textunderscore hpo} should be set to {\it True} in the {\it uq}-section of the configuration file with the RBFs chosen as the surrogate model in the {\it hpo}-section.} 

The first approach can also be adapted to incorporate the computed variances $\boldsymbol{V}_{\text{model}}$ by appending them into the outer loss function $\ell_1$. 
In the HYPPO software, a setting is available to add a regularization term to the outer objective to create a regulated loss function $\ell_{\text{reg}}$,
\begin{multline}\label{eq:ell_reg}
\ell_{\text{reg}}\left(\boldsymbol{\theta}; \bw^*, \mathcal{D}_{\text{val}}\right) =\ell_1\left(\boldsymbol{\theta}; \bw^*, \mathcal{D}_{\text{val}}\right)\\ + \gamma \sum_{d=1}^{D}g\left(\boldsymbol{V}_{\text{model}}\left( \mathbf{x}^d \right)\right),
\end{multline}
where $g: \mathbb{R}^d \rightarrow \mathbb{R}^+$ and $\gamma > 0$. 
Thus, choosing $\gamma$ to be large relative to the first term places a substantial emphasis on minimizing the performance variability during the surrogate modeling. 
A user could also tailor $g$ to magnify or dampen the variance to varying degrees or even for particular ranges of inputs by defining $g$ in a piece-wise fashion, e.g. $g(\mathbf{x}) = ||\max(0, \mathbf{x})||$. 
If this setting is applied, then the value of $\ell_{\text{reg}}$ is used by the surrogate models while the confidence interval of $\ell_1$ is returned during the HPO as previously noted.

\begin{figure}[ht]
\centering
\includegraphics[width=\linewidth]{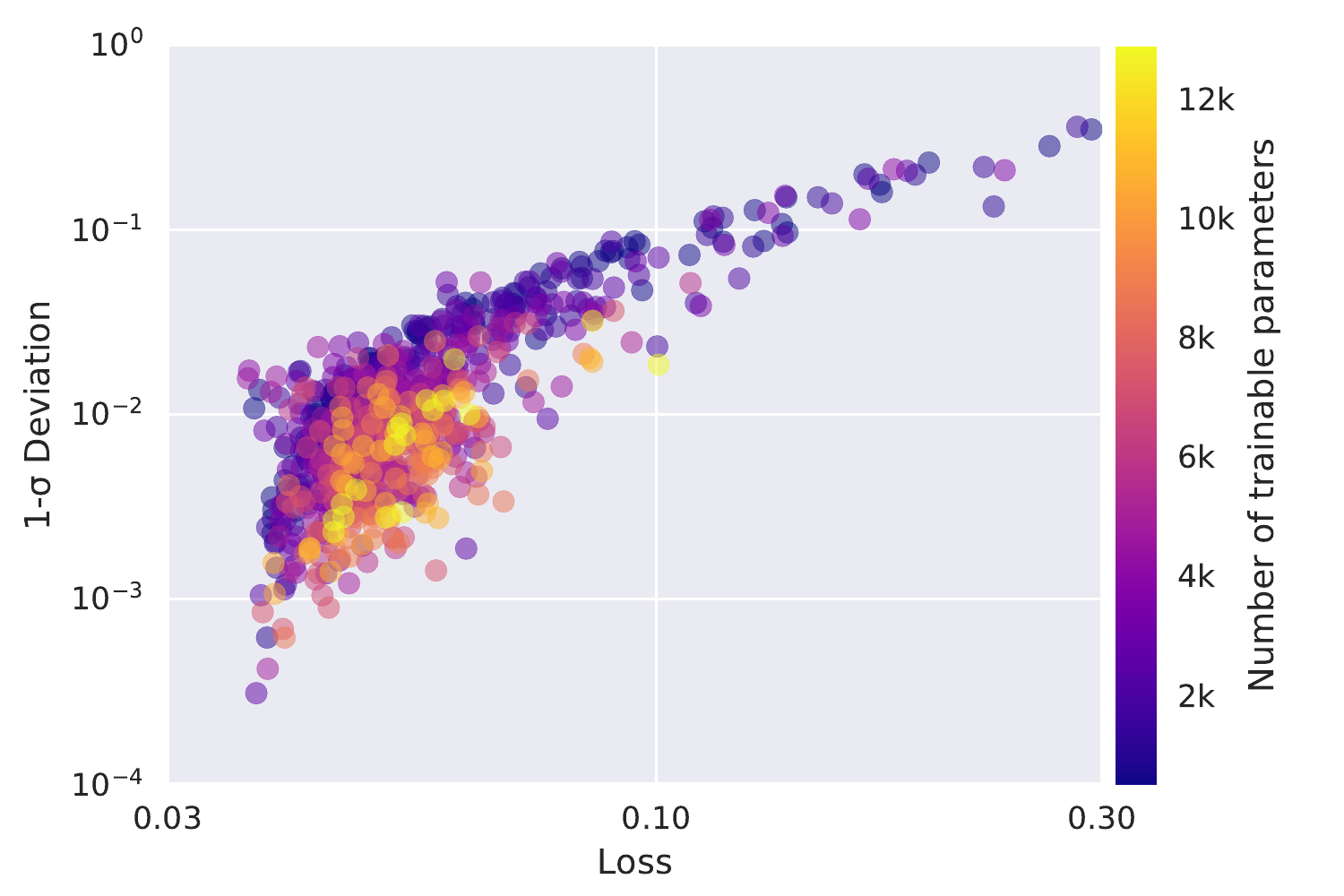}
\caption{Distributions of the loss, standard deviation and total number of trainable parameters across 825 trained multi-layer perceptron (MLP) models with different hyperparameter sets. The models were used to do time series prediction of the daily temperature in Melbourne, Australia. While complex architectures, i.e. high number of trainable parameters, are found to be clustered, low-complexity models can be identified in the low-loss, low-uncertainty region, making them suitable model solutions.}
\label{user_results}
\end{figure}

In Fig. \ref{user_results}, we show one typical result provided by HYPPO to the user, showing the distribution of trained model according to their computed loss and confidence limit. Along with the total number of trainable parameters for each model, which characterizes the complexity of the models' architecture, this graph provides insightful information that can help users decide which model is most efficient and suitable for their problem. A combination of low uncertainty, low loss and small number of trainable parameters can be considered as an optimal choice of model.

\subsection*{Feature 2: Surrogate Modeling}
Our HPO software contains two different types of surrogate models, namely RBFs and GPs, and corresponding iterative sample strategies.
The HPO method is, however, general enough and can easily be extended to other surrogate models. 
The RBF surrogate model is of the form,
\begin{equation}\label{eqn:rbf}
  m_{\text{RBF}}(\boldsymbol{\theta}) = \sum_{j=1}^{n} \lambda_j \phi\left(|| \boldsymbol{\theta} - \boldsymbol{\theta}_j ||_2\right) + p(\boldsymbol{\theta}),
\end{equation}
\noindent where $\boldsymbol{\theta}_j$ are the sets of hyperparameters for which the objective function has previously been evaluated, $\|\cdot\|_2$ denotes the Euclidean norm, $\phi(r) = r^3$, and $p(\boldsymbol{\theta}) = \beta_0 + \boldsymbol{\beta}^T \boldsymbol{\theta}$ is a polynomial tail whose form depends on the choice for $\phi(\cdot)$. 
A system of equations (see Eq.~6 in \cite{Muller2020}), dependent on the value of $\ell_1$ (or $\ell_\text{reg}$) at the $\boldsymbol{\theta}_j$'s, is solved to compute $\lambda$, $\beta_0$ and $\boldsymbol{\beta}$. 
When using the RBF ensemble approach as described in the previous section, the various values sampled from the confidence intervals are used as the right-hand side of Eq.~6 in~\cite{Muller2020} thus generating multiple response surfaces. 

When making iterative sampling decisions with the RBF model, we follow the ideas in~\cite{Regis2007b}. We generate a large set of candidate points by perturbing the best point found so far and by randomly selecting points from the search space, making sure we satisfy the integer constraints. 
For each candidate point, we use the RBF to predict its function value (e.g., loss) and we also compute the distance of each candidate point to all previously evaluated hyperparameters. 
A weighted sum of these two criteria is then used to select the best candidate point at which the next expensive evaluation (model training, MC dropout) is performed. 
The weights cycle through a predefined pattern to enable a repeated transition between local and global search.

The second type of surrogate model implemented in our software is GPs. 
GPs have the advantage that in addition to providing a prediction of the function value, they also return an uncertainty estimate of the prediction. When using a GP, we treat the function as a realization of a stochastic process:
\begin{equation}
    m_\text{GP}(\boldsymbol{\theta}) = \nu +Z(\boldsymbol{\theta}),
\end{equation}
where $\nu$ represents the mean of the stochastic process and $Z(\boldsymbol{\theta})\sim \mathcal{N}(0,s^2)$. 
Therefore, the term $Z(\boldsymbol{\theta})$ represents a deviation from the mean $\nu$. 
It is assumed that there exists a correlation between the errors and that it depends on the distance between the hyperparameters.
For an unsampled set of hyperparameters, the GP prediction represents the realization of a random variable with mean $\nu$ and variance $s^2$. 
Further details on how to compute the mean and variance can be found in~ \cite[Eqs.~7-13]{Muller2020}. 
When using the GP surrogate model, we maximize the  expected improvement~\cite{Jones1998} auxiliary function using a genetic algorithm that can handle the integer constraints. 
This sampling strategy uses the GP-predicted value of the loss $\ell_1$ and the corresponding GP-predicted uncertainty to balance a local and a global search for improved solutions in the hyperparameter space. 
The process of utilizing MC dropout with GP remains unchanged compared to the implementation with RBF. The only difference is that no ensemble approach, as detailed for the RBFs, is needed as the GP provides the prediction mean and standard deviation.

\begin{figure}[t]
\centering
\includegraphics[width=\linewidth]{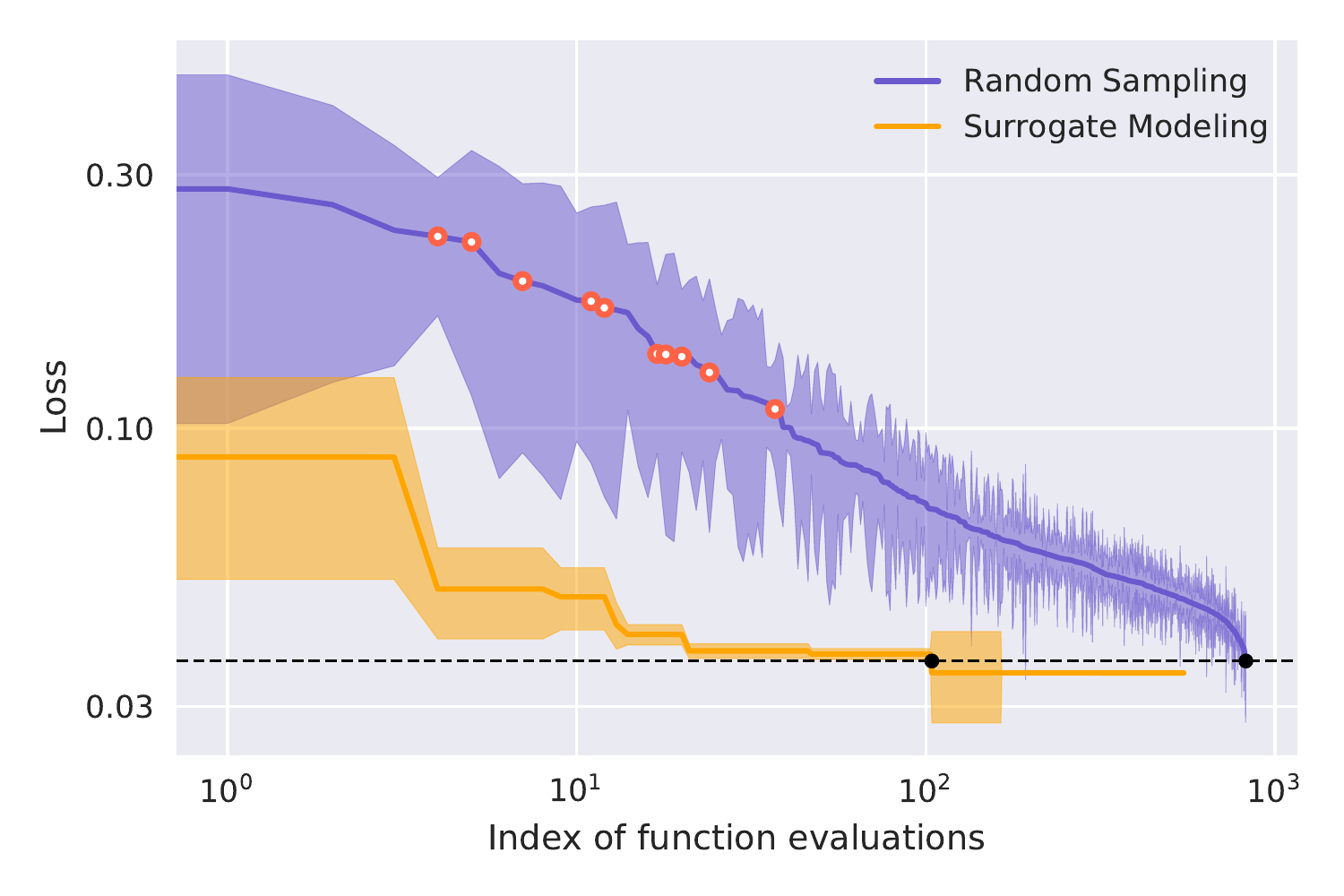}
\caption{Convergence plot showing the loss value and its 1-$\sigma$ deviation for all function evaluations. In purple, we show the sorted loss values for a random set of 825 hyperparameter samples generated using low-discrepancy sampling. From that set, 10 evaluations with high losses were selected (red points) and used as initial evaluations for surrogate modeling. In orange, we show the best error loss as the surrogate modeling is iterating and were able to decrease by an order of magnitude the amount of evaluations required to find the most optimal region in the hyperparameter space.}
\label{fig:compare}
\end{figure}

In Fig.\,\ref{fig:compare}, we demonstrate the effectiveness of using surrogate modeling and compare the increase of convergence speed to reach the optimal point in the hyperparameter space, where the lowest loss model can be achieved with an order of magnitude fewer iterations.

\begin{figure}[t]
\centering
\includegraphics[width=\linewidth]{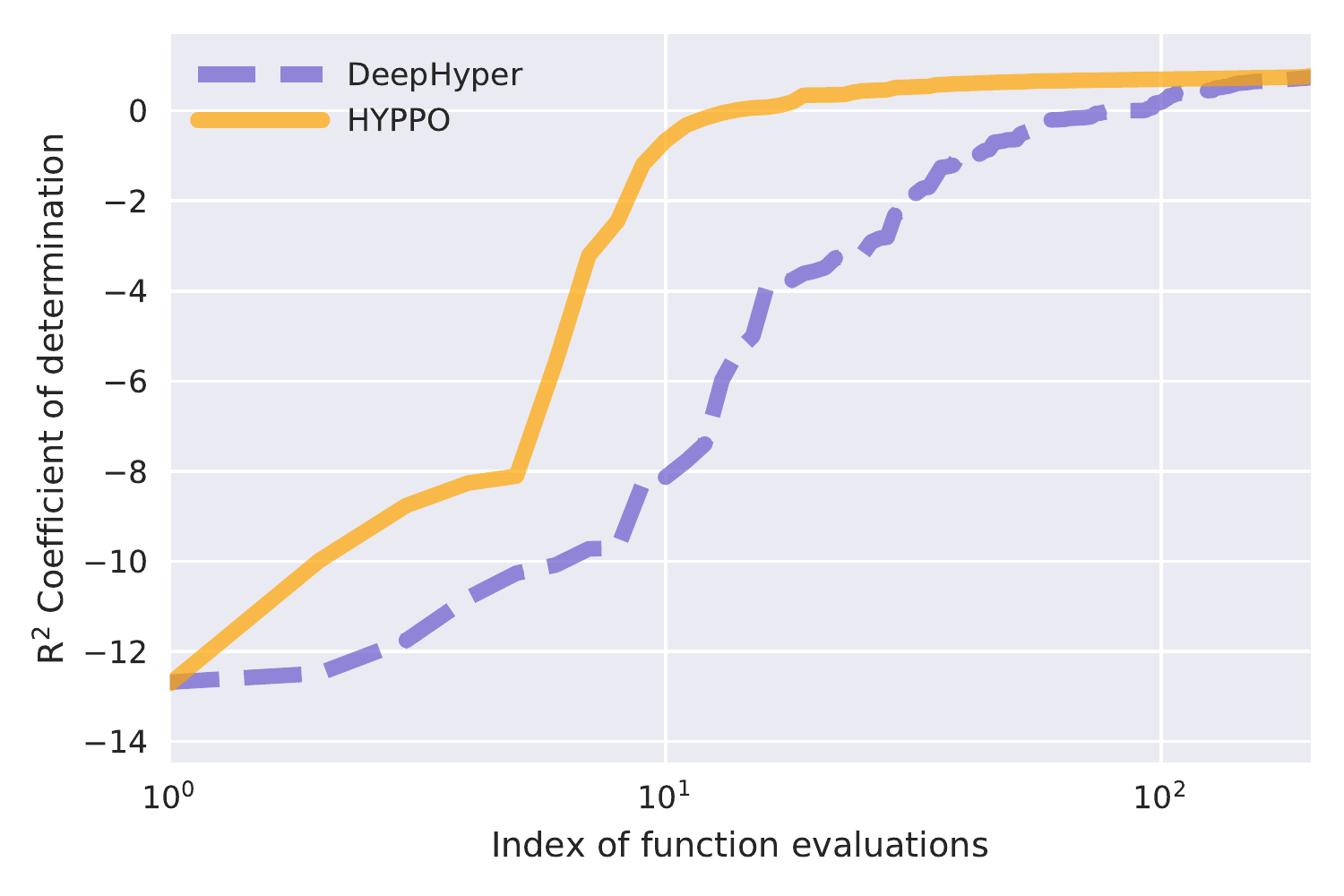}
\caption{Comparison of convergence plot between our HYPPO software and the DeepHyper HPO library. We used the polynomial fit problem provided in the DeepHyper's online documentation, allowing  200 iterations. In this example, the $R^2$ value (also known as "Coefficient of determination") is the metric being  maximized.}
\label{fig:deephyper}
\end{figure}

The improvement in HPO performance with HYPPO from a different library, e.g. DeepHyper, is shown in Fig.\,\ref{fig:deephyper}. To do the comparison, we used the polynomial fit problem provided in  DeepHyper's online documentation\footnote{\url{https://deephyper.readthedocs.io/en/latest/tutorials/hps_dl_basic.html}} and we extended the complexity of the problem by increasing the number of hyperparameters to be optimized to six: (1) number of nodes per layer, (2) number of layers, (3) dropout rate, (4) learning rate, (5) epochs, and (6) batch size. We applied HPO with both HYPPO and DeepHyper libraries independently and evaluated the high\-dimensional hyperparameter space at a total of 200 iterations with each method. Ten initial evaluations were used in HYPPO to build the initial surrogate model. Our analysis shows that HYPPO  and DeepHyper both find eventually models with the same performance, but HYPPO finds better hyperparameters faster (fewer iterations) than DeepHyper, adding further  motivation to using HYPPO.

\subsection*{Feature 3: Asynchronous Nested Parallelism}
One of the challenges in carrying out HPO for machine learning applications is the computational requirement needed to train multiple sets of hyperparameters. In order to address this problem, the HYPPO software handles parallelization across multiple hyperparameter evaluations and enables distributed training for each evaluation. 
Here, we describe how the nested parallelism feature is being implemented and used to maximize efficiency across all allocated resources.

\subsubsection{Nested parallelism}

\begin{figure}[t]
\begin{center}
\includegraphics[width=\linewidth]{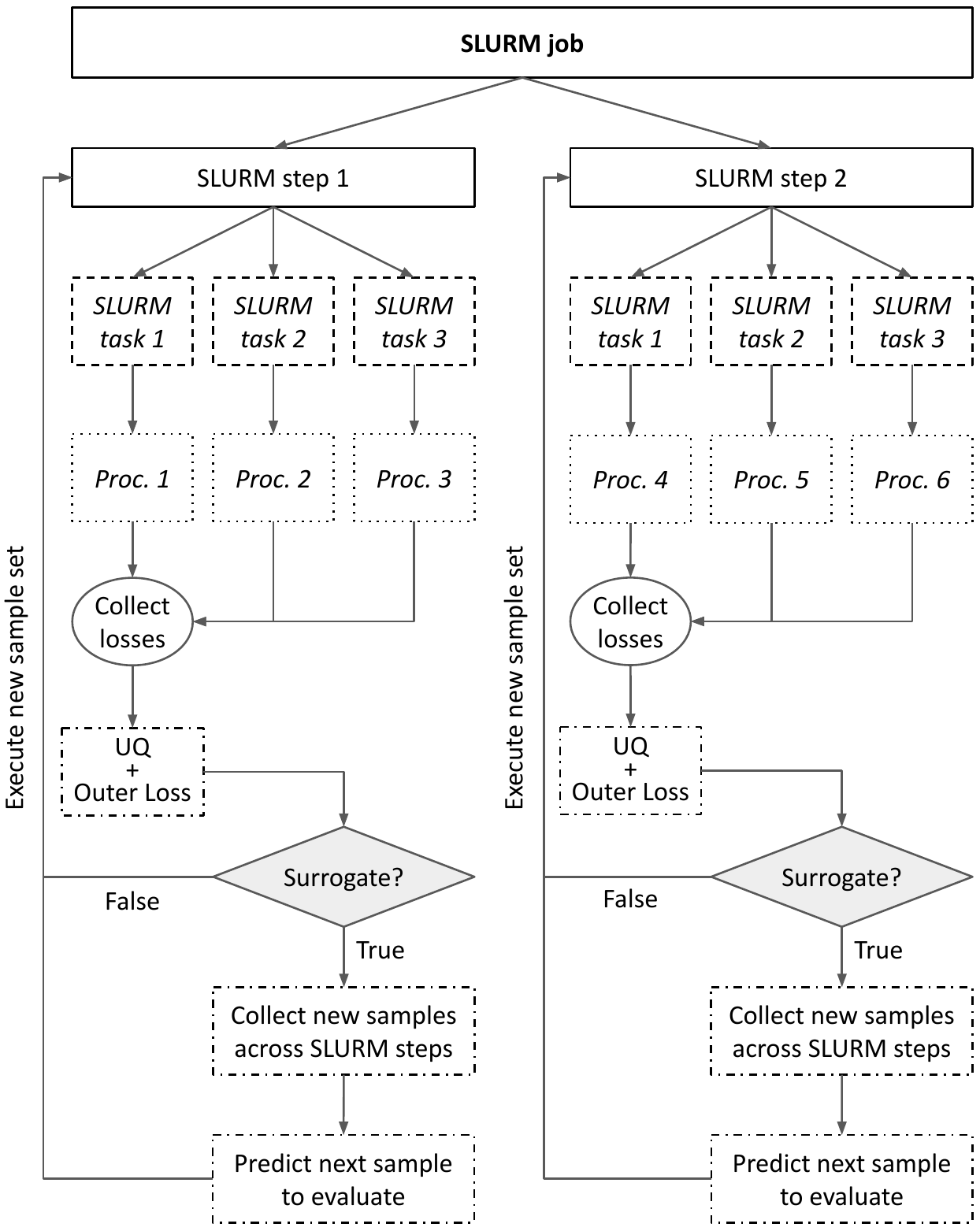}
\caption{Diagram illustrating the nested parallelization process across 6 processors using 2 parallel steps and 3 processors per step. A single processor is used per SLURM task. If the user requests uncertainty quantification to be performed on the trained models, the outer loss will be provided along with its confidence interval. When surrogate modeling is performed, the algorithm will gather newly evaluated sample sets across all parallel steps and make a prediction for the next set of hyperparameters to evaluate.}
\label{fig:parallel}
\end{center}
\end{figure}

In Fig. \ref{fig:parallel}, we show how we set up the parallelization scheme in the software. 
The HYPPO software uses GNU parallel~\cite{tange_ole_2018_1146014} to do distributed training of multiple models in parallel. The program can automatically generate a SLURM script using the number of SLURM steps to be executed in parallel (that is, the number of individual \texttt{srun} instances) and the number of SLURM tasks to be executed in each step, as defined by the user in the input configuration file. Since a single processor (either GPU or CPU, the user decides) will be assigned for each task, the total number of processors used in a SLURM job is therefore equal to the number of steps times the number of tasks. 
For instance, proper SLURM directives for a setup with 2 \texttt{srun} instances running in parallel across 3 GPUs for each step will read as follows:

\begin{verbatim}
    #SBATCH --ntasks 6
    #SBATCH --gpus-per-task 1
\end{verbatim}

In this case, \texttt{--ntasks} is equal to the total number of processors to be allocated for the job. 
Each of the 2 SLURM steps can then be executed in parallel using GNU parallel with a \texttt{--jobs} command set to 2. 
Finally, in order to avoid individual \texttt{srun} steps to be executed on the same GPUs, we use the \texttt{--exclusive} command for the \texttt{srun} call which will ensure that separate processors will be dedicated to each job step.

Once the program starts, workers will be initialized according to the SLURM settings. 
If the used machine learning library is PyTorch, the program will do the initialization using the \texttt{torch.distributed} package. 
On the other hand, if the Tensorflow library is used, the Horovod package will be used to initialize the workers. 
While the software is flexible enough to work with external training modules, the library used in the external package needs to be specified in the input configuration file so that the program can initialize the workers accordingly.

Using the SLURM environment variables, we can then loop over the randomly generated hyperparameter sets, ensuring that each step executes a different set of hyperparameters. 
This can be achieved using Python's slicing feature where the subsequence is defined using the step ID and the total number of steps in the job. 
When multiple processors are requested for each evaluation, the loss for the outer objective function will always be calculated in the first processor (e.g., GPU0 if GPU processors are used) and the value will be recorded in its corresponding log file. 
In the meantime, the remaining processors will wait for the task to be completed by searching for the recorded value in the first processor's log file. 

\subsubsection{Data / Trial Parallelization}\label{sec:parallelization}

As quantifying the uncertainty becomes more accurate with an increasing number of trials (i.e., repeatedly training the same architecture), the HYPPO software offers the possibility, in a single SLURM step, to parallelize the trials instead of the data to be trained on.
Thus, in the example in Figure~\ref{fig:parallel}, the SLURM job corresponds to performing HPO, SLURM steps 1 and 2 correspond to two different sets of hyperparameters to be evaluated, and SLURM tasks 1-3 correspond to performing three separate trainings  of each hyperparameter set. 
For example, suppose for a specific hyperparameter set, the model is to be trained nine times.
In that case, each GPU will execute three consecutive independent trainings, and slicing will be applied to the trial indices using the extracted MPI rank and size values to ensure that the total number of trials to be computed across all GPUs in each step is equal to the number of trials requested by the user. 
On the other hand, if the user prioritizes the parallelization over the data (train in parallel),  then in the example in Figure~\ref{fig:parallel}, the SLURM job still corresponds to performing HPO, SLURM steps 1 and 2 correspond to two different sets of hyperparameters to be evaluated, and SLURM tasks 1-3 perform the parallel training using a third of the data per GPU. 
Again, if the goal was to train each hyperparameter set nine times, each GPU will execute all nine independent training trials sequentially.

%\textcolor{blue}{[Juli says: Vincent, if you dont think the above blue text is helpful, just relete and keep your paragraph below]}
%For instance, let's consider a step that has 3 allocated GPUs and a model that is to be trained 9  times. If the user decides to parallelize over trials, each GPU will execute 3 consecutive independent trainings of the model and slicing will be applied to the trial indices using the extracted MPI rank and size values to ensure that the total number of trials to be computed across all GPUs is equal to the number of trials requested by the user. If the user, on the other hand, decides to prioritize the parallelization over the data, each GPU will then execute all 9 independent training trials sequentially but using a third of the data per GPU.\\

\subsubsection{Asynchronous update}

\begin{figure}[t]
\begin{center}
\includegraphics[width=\linewidth]{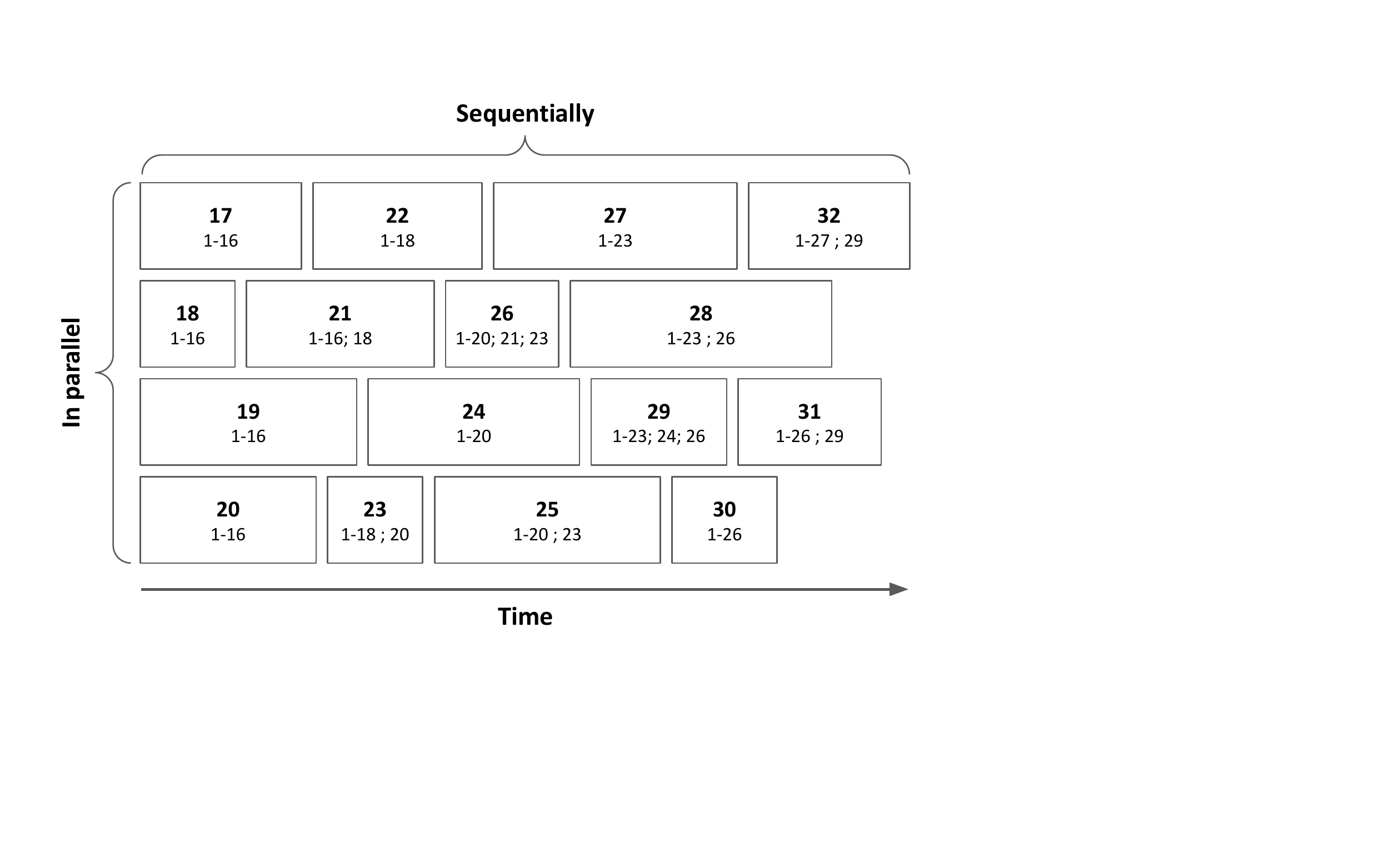}
\caption{Illustration of asynchronous surrogate modeling. Each rectangle represents an evaluation, the top number corresponds to the evaluation index and the bottom values are the indices of the evaluations that were used by the surrogate model to determine the current  hyperparameter set. In this example, 16 initial evaluations were done prior to starting the surrogate modeling and asynchronous parallelism. The model was then fitted and predicted 4 new sample sets to evaluate (evaluations 17 through 20 shown in bold). Once the first evaluation completed, i.e. evaluation 18 in this example, a surrogate model was refitted using the newly computed evaluation (in addition to the first 16 evaluations) and a new set of hyperparameters to evaluate was predicted (here, evaluation 21).}
\label{fig:async}
\end{center}
\end{figure}

The surrogate model-based optimization is updated every time a new input-output data pair is obtained during the surrogate model-based optimization.
This process is usually done sequentially (one evaluation per iteration and update of the  model) or synchronously parallel (multiple evaluations simultaneously and update the model only after all evaluations are completed). 
This may require many iterations before the surrogate model adequately represent the loss function over the entire hyperparameter space. Moreover, in HPO, each hyperparameter evaluation may require a different amount of time because architectures of different sizes have different numbers of weights and biases that the training process optimizes.
Using the aforementioned nested parallelization feature, we can use this approach to predict not one but multiple sets of samples to be evaluated asynchronously in parallel. 
The surrogate model is then updated each time a hyperparameter set has been evaluated.

Nevertheless, unlike the initial evaluations that are performed independent of each other, the surrogate modeling process requires communication between the SLURM steps in order to retrieve newly evaluated hyperparameter sets from other \texttt{srun} instances and update the model. In Fig.~\ref{fig:async} we illustrate how evaluations can complete at different times during the surrogate modeling process. After each completed evaluation, the HYPPO software reads through all the log files generated and constantly updated by each processor to search for newly computed sample sets. Once the search is complete, the surrogate model is updated and a new sample set is computed.

\section{Case Study: CT Image Reconstruction}\label{sec:CT}
% \subsection{Science Problem}

This section describes a relevant scientific application from computed tomography (CT) and demonstrates the impact of our HPO process on the results. 
CT is a three-dimensional imaging technique that measures a series of two-dimensional projections of an object rotated on a fixed axis relative to the direction of an X-ray beam. 
The collection of projections from different angles at the same cross-sectional slice of the object is called a 2D sinogram, which is the input to a reconstruction algorithm. 
The final 3D reconstructed volume is called a tomogram, which is assembled from the independent reconstructions of each measured sinogram. 
Tomographic imaging is used in various fields, including biology~\cite{wise_micro-computed_2013}, medicine~\cite{rubin_computed_2014}, materials science~\cite{garcea_x-ray_2018}, and geoscience~\cite{cnudde_high-resolution_2013}, allowing for novel observations that enable enhanced structural analysis of subjects of interest.

% \begin{figure}[h]
%     \centering
%     \includegraphics[width=\linewidth]{imgs/example_dataset_CT.png}
%     \caption{Test data examples for the CT reconstruction problem.}
%     \label{fig:example_dataset_CT}
% \end{figure}

High-quality tomographic reconstruction generally requires measuring projections at many angles. 
However, collecting sparsely sampled CT data (i.e., sparse distribution of angle measurements) has the benefit of exposing the subject to less radiation and improving temporal resolution. 
In sparse angle CT, where few angles are available in the measured sinograms, computing the reconstruction is a severely ill-posed inverse problem. 
When using standard algorithms, the reconstructions will be noisy and obstructed with streak-like artifacts. 
In recent years, a class of algorithms using deep learning has come into prominence~\cite{parkinson_machine_2017, pelt_improving_2018, ayyagari_image_2018, huang_investigations_2018, bazrafkan_deep_2019, fu_hierarchical_2019, he_radon_2019}. 
Such approaches show superior performance compared to standard algorithms by using a trained neural network to solve the inverse problem of sparse angle CT reconstruction. 

Here, we use HPO to optimize a deep neural network architecture for performing sinogram inpainting, an approach that has found success in other works~\cite{lee_view-interpolation_2017,li_sinogram_2019}. 
The missing angles of a sparsely sampled sinogram are filled in by a trained neural network, after which the completed sinogram can be reconstructed using any standard algorithm.

\subsection{Data and Architecture}\label{sec:data}

\begin{figure}[h]
    \centering
    \includegraphics[width=\linewidth]{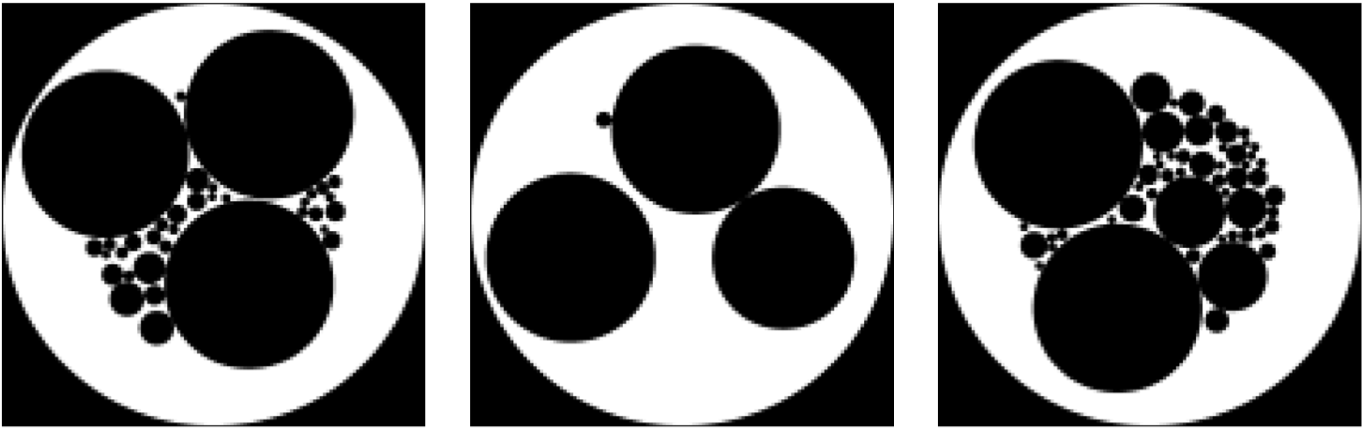}
    \caption{Test data examples for the CT reconstruction problem.}
    \label{fig:example_dataset_CT}
\end{figure}

We use a simulated dataset created with \textit{XDesign}, a Python package for generating X-ray imaging phantoms~\cite{ching2017xdesign}. 
The dataset comprises $17500$ images of $128 \times 128$ pixels ($13500$ training, $1500$ validation, $2500$ test examples) of circles of various sizes, emulating the different feature scales present in experimental data as shown in~Fig.~\ref{fig:example_dataset_CT}. 
We use TomoPy~\cite{gursoy2014tomopy}, a Python package for tomographic image reconstruction and analysis,
to generate sinograms of these images with $20$ angles. To emulate sparse angle CT, every other angle is removed from the sinogram and Poisson noise is added.

The neural network architecture used in this problem takes the form of a U-Net, a type of CNN that has shown success in biomedical image processing problems~\cite{ronneberger2015u}. 
The input of the U-Net is the sparse angle sinogram, and the desired output is the completed sinogram. 
Our variation of the U-Net architecture consists of an equivalent number of downsampling and upsampling blocks. 
Within each block, several intermediate convolution layers preserve the size of the input, and a final convolution layer increases the number of feature maps (channels). 
We selected eight hyperparameters for HPO, enumerated in Table~\ref{table:params}. 
HPO aims to minimize the mean squared error between the actual and predicted sinograms of the validation dataset.

\begin{table}[h!]
\centering
\caption{A total of eight hyperparameters were selected for hyperparameter optimization: (1) number of output feature maps of the initial block, (2) multiplier for the number of feature maps in each subsequent block, (3) number of blocks, (4) intermediate layers, (5) convolutional kernel size and (6) stride of the final layer in each block, (7) dropout probability and (8) intermediate layer kernel size. The first eight rows show the value of each hyperparameter and the last four rows contain the training results and image metric quantities averaged over the test dataset. Columns (a) and (d) are the results of the network trained with the minimum and maximum values for each hyperparameter, respectively. Columns (b) and (c) are the results of training the neural network with the best and worst hyperparameter values sampled by  HYPPO, respectively.}
\begin{tabular}{l || l l l l} 
 \toprule
 \textbf{Parameters} & \textbf{(a)} & \textbf{(b)} & \textbf{(c)} & \textbf{(d)}\\
 \midrule
 \textbf{(1)} & $8$ & $9$ & $10$ & $12$\\
 \textbf{(2)} & $1.0$ & $1.0$ & $1.2$ & $1.4$ \\
 \textbf{(3)} & $2$ & $2$ & $3$ & $4$ \\
 \textbf{(4)} & $1$ & $1$ & $4$ & $4$ \\
 \textbf{(5)} & $2$ & $3$ & $4$ & $5$ \\
 \textbf{(6)} & $1$ & $1$ & $2$ & $2$ \\
 \textbf{(7)} & $0.00$ & $0.01$ & $0.08$ & $0.1$ \\
 \textbf{(8)} & $2$ & $3$ & $5$ & $5$ \\
 \textbf{MSE} & $2.98$E$-5$ & $3.39$E$-5$ & $1.56$E$-3$ & $2.46$E$-3$ \\
 \textbf{PSNR} & 35.2 & 34.6 & 18.7 & 16.4 \\
 \textbf{SSIM} & 0.965 & 0.988 & 0.970 & 0.955 \\
 \bottomrule
\end{tabular}
\label{table:params}
\end{table}

The resulting sinogram from the trained U-Net is reconstructed by TomoPy using the Simultaneous Iterative Reconstruction Technique (SIRT) \cite{GILBERT1972105}, which has shown the ability to produce accurate reconstructions given incomplete datasets. SIRT iteratively minimizes the error between the measured projections and the projections calculated from the reconstruction at the current iteration $k$. 
The average error is then backprojected to refine the reconstruction at each iteration. 
For an inverse problem $\textbf{Ax} = \textbf{b}$, the SIRT update equation is: \[\textbf{x}_{k+1} = \textbf{x}_k + \textbf{CA}^\top\textbf{R}\left(\textbf{b}-\textbf{Ax}_k\right),\] where $\textbf{C}$ and $\textbf{R}$ are diagonal matrices that contain the inverse of the sum of the columns and rows, respectively, of $\textbf{A}$. 

\subsection{Results}

\begin{figure}[h]
    \centering
    \includegraphics[width=\linewidth]{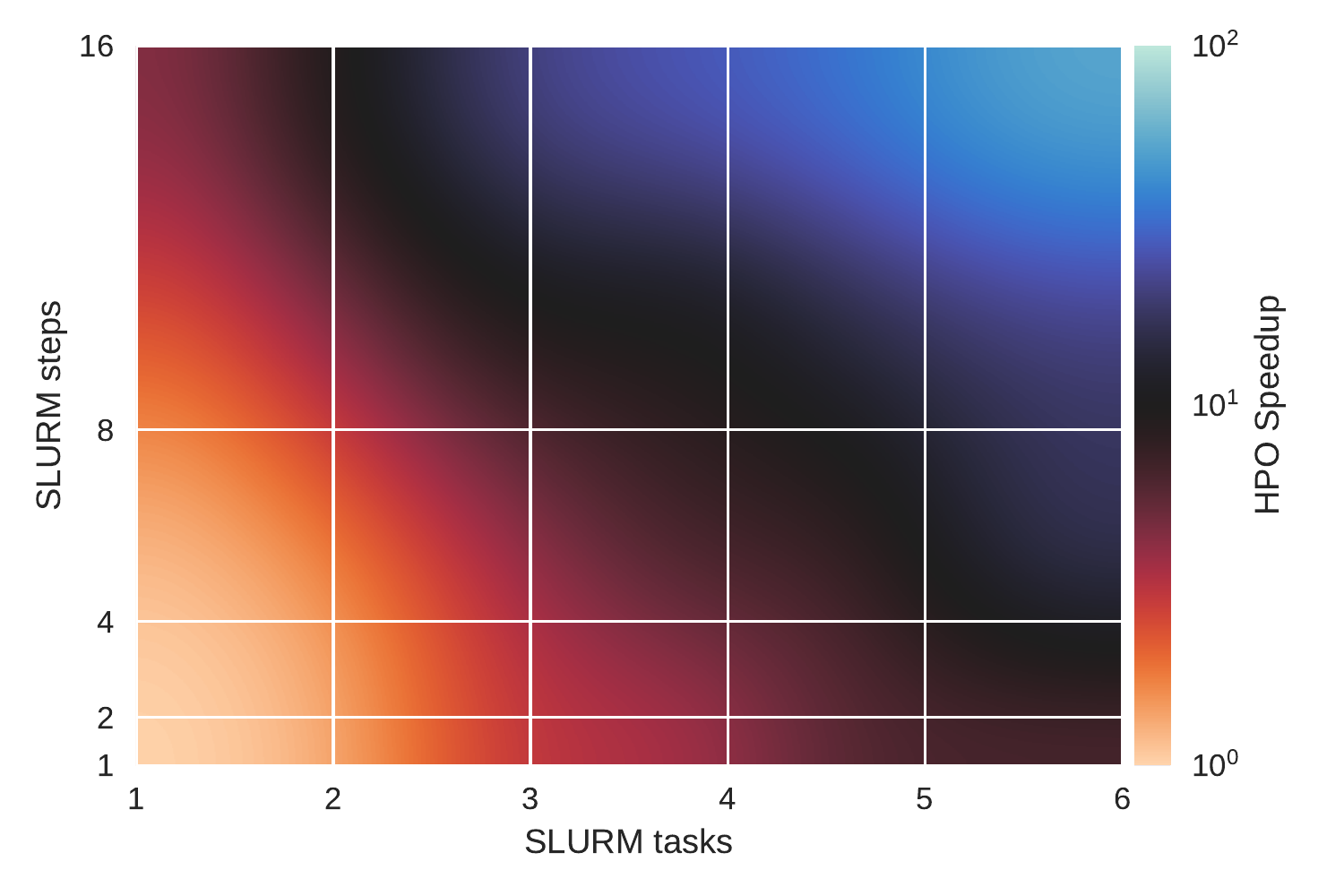}
    \caption{Job speedup as a function of the number of SLURM steps and number of SLURM tasks is shown for evaluating 50 different hyperparameter sets and five trials for each. Up to 96 GPU processors from the NERSC/Cori cluster were used to evaluate performance and scalability of our software. Lower speedup values correspond to better performance.}
    \label{fig:speedup}
\end{figure}

We analyze two aspects of applying HPO to the case study of CT image reconstruction: 1) job speedup as a function of the number of SLURM steps and SLURM tasks for the specific configuration of running the evaluation operation (see Sec. \ref{sec:parallelization}) and 2) improvement due to optimizing the hyperparameters. HYPPO is executed on the Cori supercomputer, housed at the National Energy Research Scientific Computing Center at Lawrence Berkeley National Laboratory (NERSC). Cori is a Cray XC40 system comprised of 2,388 nodes each containing two 2.3 GHz 16-core Intel Haswell processors and 128 GB DDR4 2133 MHz memory, and 9,688 nodes each containing a single 68-core 1.4 GHz Intel Xeon Phi 7250 (Knights Landing) processor and 96 GB DDR4 2400 GHz memory. Cori also provides 18 GPU nodes, where each node contains two sockets of 20-core Intel Xeon Gold 6148 2.40 GHz, 384 GB DDR4 memory and 8 NVIDIA V100 GPUs (each with 16 GB HBM2 memory). We use the Haswell processor and the GPU nodes for the results presented in this section. 

\begin{figure}[h!]
    \centering
    \includegraphics[width=0.98\linewidth]{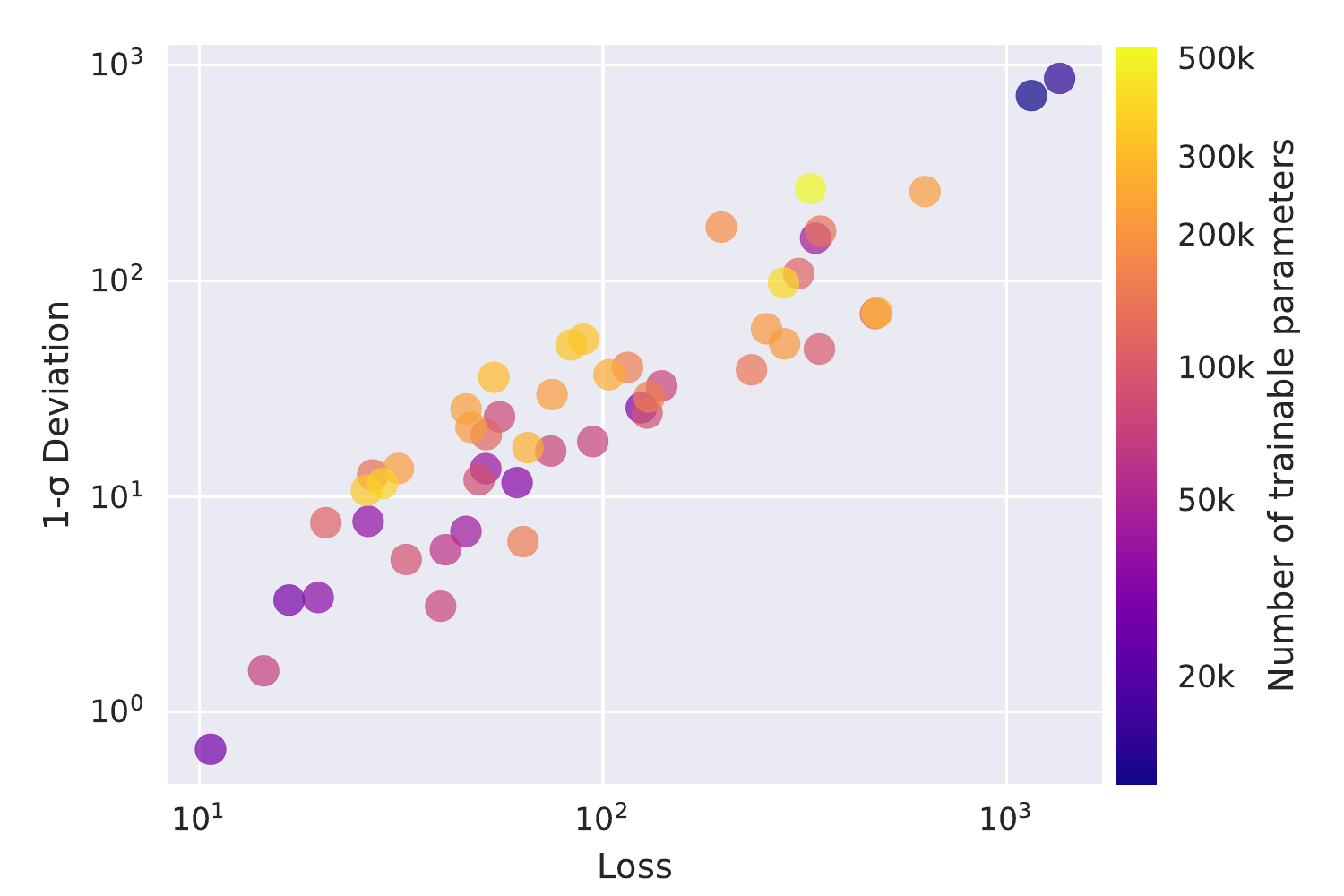}
    \caption{Scatter plot of median loss value versus median absolute deviation. The loss and median absolute deviation are computed over 50 training trials for each of 50 hyperparameter evaluations. In the bottom left region, a simple architecture (less than 50k trainable parameters) can be found which provides an accurate (low loss) and efficient (low variability) solution.}
    \label{fig:ct_scatterplot}
\end{figure}

\begin{figure}[h!]
    \centering
    \includegraphics[width=\linewidth]{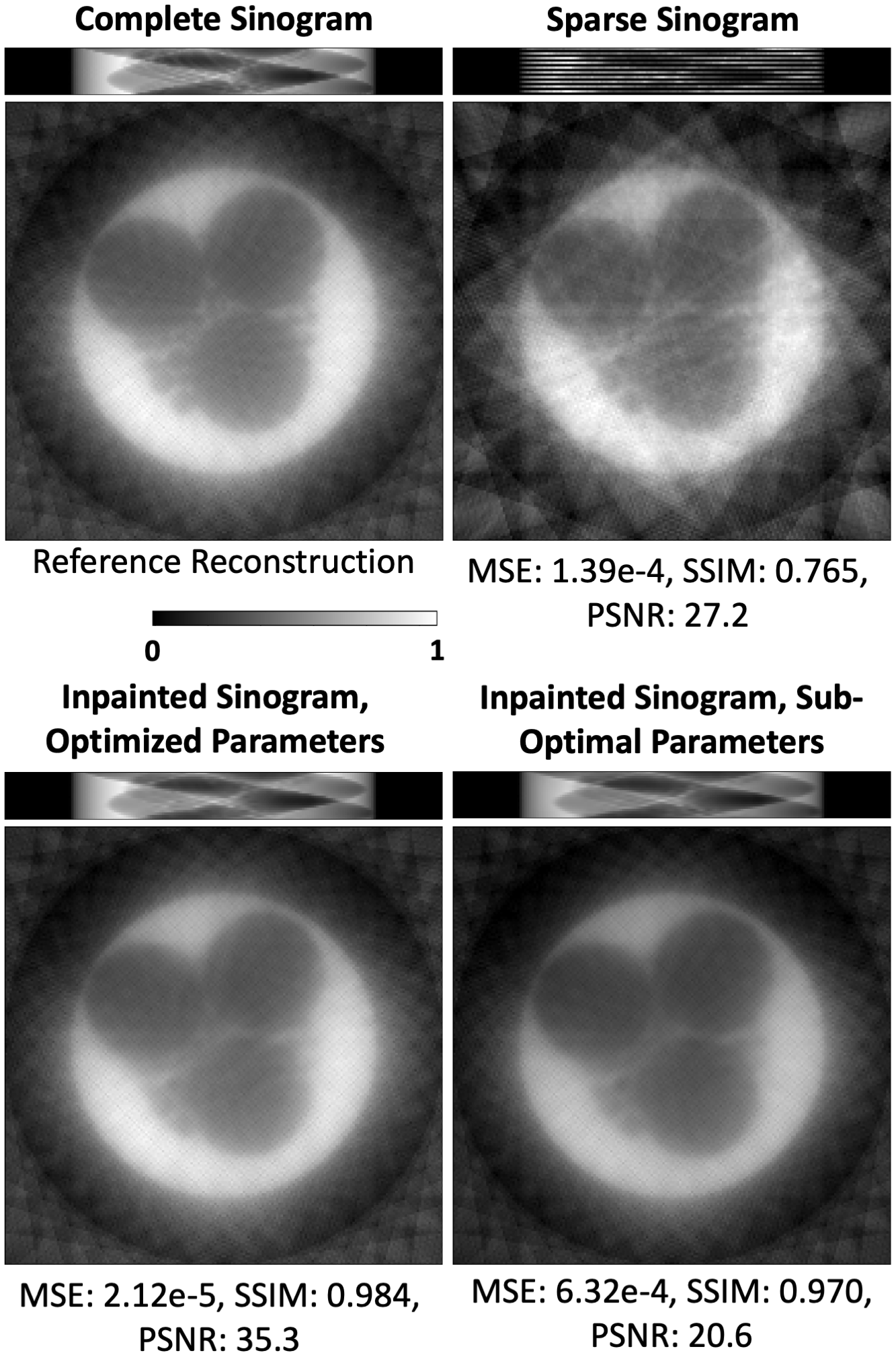}
    \caption{Comparison of the complete sinogram with the sparse and inpainted (neural network output) sinograms for a test example. The reconstruction for each sinogram is performed using the SIRT algorithm, described in Section \ref{sec:data}. Error metrics of per-pixel mean squared error (MSE, lower is better), structural similarity (SSIM, larger is better), and peak signal-to-noise ratio (PSNR, larger is better) compared to the reference reconstruction, showing the improvement of using the inpainted sinogram with optimized network hyperparameters over the sparse sinogram and inpainted sinogram with sub-optimal network hyperparameters.}
    \label{fig:sinogram_output}
\end{figure}

The first experiment consists of running 50 different hyperparameter evaluations with five trials for each evaluation. Fig.~\ref{fig:speedup} reports the maximum speedup resulting from this experiment across each SLURM step and SLURM task. 
The neural network is trained for $300$ iterations using the entire training dataset of $13500$ images. 
We observe an improvement of two orders of magnitude in speedup between the combination of 1 SLURM task/1 SLURM step and 6 SLURM tasks/16 SLURM steps.

Fig. \ref{fig:ct_scatterplot} shows the outer loss function value plotted against the median absolute deviation. Here, we evaluated each hyperparameter set $50$ times to compute the median outer loss function value and standard deviation. An outer loss value of 24.81 is achieved within four iterations when performing Gaussian process surrogate modeling, which may not be guaranteed by random sampling alone, emphasizing the importance of HPO in this case study.

\begin{figure}[h!]
    \centering
    \includegraphics[width=\linewidth]{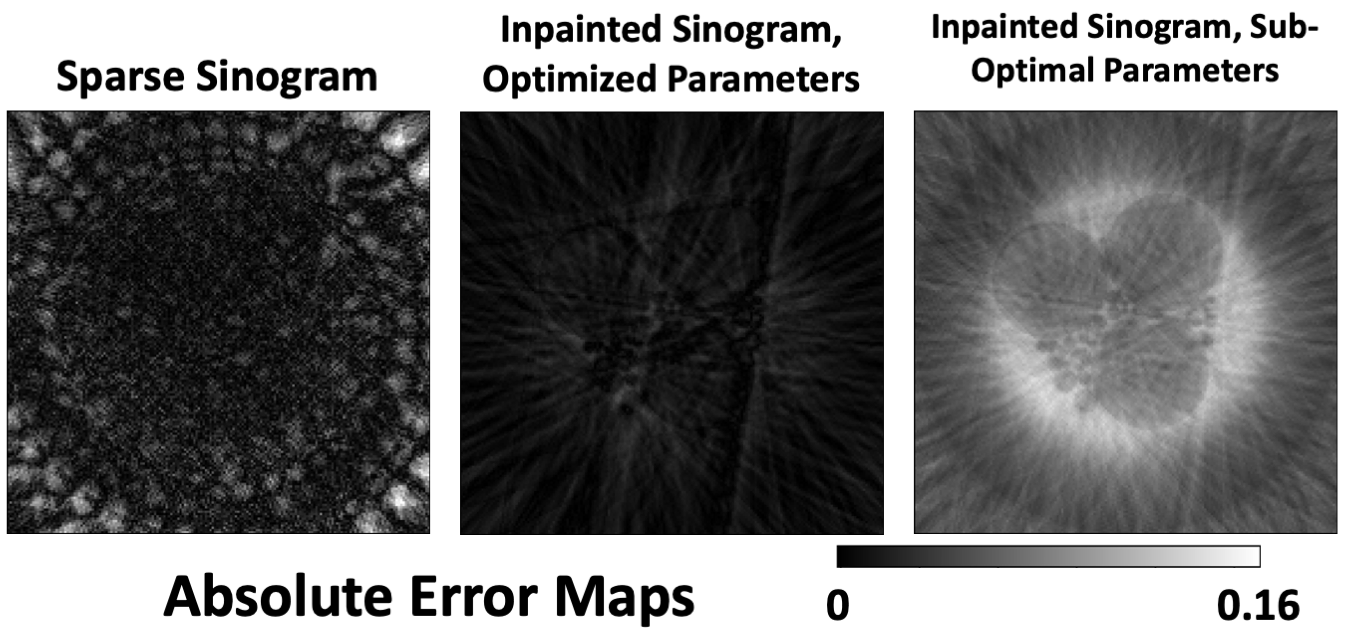}
    \caption{The absolute error maps for each example in Fig.~\ref{fig:sinogram_output}. The baseline comparison for each error map is the complete sinogram.}
    \label{fig:sinogram_output_error}
\end{figure}

We also assess the performance of the U-Net after running HPO, comparing the quality of the reconstruction obtained with the neural network output to the original (complete) sinogram. 
The optimal network hyperparameters found with HPO show improved reconstructions compared to other    hyperparameters sampled during HPO  (see Figs.~\ref{fig:sinogram_output} and~\ref{fig:sinogram_output_error}, and Table~\ref{table:params}) demonstrating the importance of HPO to this case study. 
We also show the results of training the network for  values at  the lower and upper bounds (set by the user) of all eight hyperparameters, which  define the hyperparameter search space.
For the results in Figs.~\ref{fig:sinogram_output} and~\ref{fig:sinogram_output_error}, and Table~\ref{table:params}, each neural network architecture was trained for $100000$ iterations. 
We use the structural similarity index measure (SSIM) to quantify how similar the sparse and inpainted sinogram reconstructions are to the complete sinogram reconstruction. 
SSIM values range between $-1$ and $1$, with values closer to $1$ indicating very similar structures between the two images being compared. 
SSIM, unlike MSE and PSNR, does not measure absolute error, allowing structural differences to be characterized. %(\textcolor{red}{I'm going to reword this sentence, it's confusing -- juli says: add range of SSIM (-1,1) or (0,1) or whatever it was and say that values close to 1 indicate better performance}). 

% \begin{table}
% \centering
% \caption{(a) Lower bounds, (b) optimal values, (c) sub-optimal values, and (d) upper bounds for the eight optimized hyperparameters.}
% \begin{tabular}{||c c c c c c c c c ||} 
%  \hline
%   Params & (1) & (2) & (3) & (4) & (5) & (6) & (7) & (8) \\ [0.5ex] 
%  \hline\hline
%   (a) & 8 & 1.0 & 2 & 1 & 2 & 1 & 0.00 & 2 \\ 
%  \hline
%  (b) & 9 & 1.0 & 2 & 1 & 3 & 1 & 0.01 & 3 \\
%  \hline
%  (c) & 10 & 1.2 & 3 & 4 & 4 & 2 & 0.08 & 5 \\
%  \hline
%  (d) & 12 & 1.4 & 4 & 4 & 5 & 2 & 0.1 & 5 \\ [0.5ex] 
%  \hline
% \end{tabular}
% \label{table:params}
% \end{table}

\section{Discussions}
In the previous sections we demonstrated different features of our HYPPO software for finding optimal hyperparameters  of neural nets. 
Although our results are promising on different applications, two aspects of our HYPPO method require  further development in order to improve its efficiency. First, a sensitivity analysis (SA) of the model's performance regarding the hyperparameters is needed. 
If we could identify the subset of hyperparameters that most impact the model's performance, we could significantly reduce the number of hyperparameter sets that need to be tried during the optimization because the search space would be smaller. 
Thus, additional savings in optimization time could be achieved.
Second, our current initial experimental design is created by randomly sampling integer values from the hyperparameter space. 
Using a type of space-filling design (e.g., a low-discrepancy sequence) instead would be preferable, and thus our initial surrogate models could be improved.  However, there are obstacles for both space-filling designs and sensitivity analysis when parameters have integer constraints, as in our problem. 
Off-the-shelf SA methods such as the ones implemented in the SALib open-source library in Python~\cite{Herman2017} only work for continuous parameters.
Low discrepancy sampling methods such as Sobol's sequences~\cite{sobol2001} generate evenly distributed points across the sample space, avoiding large clusters and gaps between the points. 
However, these methods are not easily modified for integer constraints and must be developed further. 
Simply rounding or truncating continuous values to obtain integers does not deliver the required sample characteristics for SA and sample designs to be maximally effective. 
However, it is possible to formulate and solve an integer optimization problem to achieve the desired sample distribution. 
There is also an opportunity to modify the computation of the sample points in Sobol's sequences. 
These aspects will be a future feature in our software. 

Another aspect in our HYPPO software that requires further analysis is the parameters, including initial experimental design sizes, the weights $w_T$ and $w_D$ used to balance the importance of expected performance and uncertainty, the number of times a given hyperparameter set should be trained, and how many hyperparameters should be tried in total. 
The weights $w_T$ and $w_D$ are user-defined and should reflect the user's averseness to model variability. 
The number of hyperparameters to be tried in total usually depends on the amount of available compute budget.  
The size of the initial experimental design influences how well the initial surrogate model approximates the objective function. 
% Usually larger numbers give better surrogate model performance initially, but it also means that fewer adaptively selected hyperparameters will be evaluated due to the compute budget. 
Typically, larger initial experimental designs give better surrogate model performance at first, but it also means that fewer adaptively selected hyperparameters will be evaluated due to the compute budget.
Another potentially impactful parameter is the dropout rate. Currently, a default value is used, but in the future we will include it in our hyperparameters to be optimized. 
Lastly, in this study we did not take noise in the data into account. In the future we will analyze how small variations in the training data propagate through the network and impact the predictive performance and reliability of the DL models. 
%\textcolor{blue}{Juli: talk about parameter adjustments?  $w_T = w_D = 0.5$ and $T = 30$.}

% Another aspect of further development concerns GPU memory distribution \textcolor{red}{Vincent please add a few words}

\section{Conclusions}
In this paper, we demonstrated a new method for conducting hyperparameter optimization of deep neural networks while taking into account the prediction uncertainty that arises from using stochastic optimizers for training the models. Computationally cheap surrogate models are exploited to reduce the number of expensive model trainings that have to be done. We showed the quality of the solutions found on a variety of datasets and how asynchronous nested parallelism can be exploited to significantly accelerate the time-to-solution. The HYPPO software comes with a number of features,  allowing the user to conduct simple HPO, UQ, and HPO under uncertainty. 
Model variability that arises from using stochastic optimizers when training models is rarely addressed in ML literature, but it can have a considerable effect, particularly in scientific applications for which dataset sizes are often limited. High variability of the model performance can have a significant impact on decisions being made, and these decisions should be made with confidence by using reliable models. 
% Model variability that arises from using stochastic optimizers when training models is rarely addressed in the ML literature, but it can have a large effect, in particular in scientific applications for which the size of the datasets is often limited. Large variability of the model performance can have a large impact on decisions being made, and these decisions should be made with confidence by using reliable models. 
Our software is a first step towards providing these much needed reliable and robust models. Finally, HYPPO was developed as an open-source software and will be made publicly available in the future via the \texttt{pip} Python package manager. More information about the software, including a detailed documentation can be found at \url{https://hpo-uq.gitlab.io/}.

\section*{Reproducibility}

As we strive to make this research fully transparent, step-by-step instructions are provided in the HYPPO online documentation (see link in the previous section) to allow complete reproduction of this work.

\section*{Acknowledgement}
Vincent Dumont, Mariam Kiran, and Juliane Mueller are supported by the Laboratory Directed Research and Development Program of Lawrence Berkeley National Laboratory and the Office of Advanced Scientific Computing Research under U.S. Department of Energy Contract No. DE-AC02-05CH11231. 
Vidya Ganapati and Talita Perciano were supported by the U.S. Department of Energy, Office of Science, Office of Workforce Development for Teachers and Scientists (WDTS) under the Visiting Faculty Program (VFP). 
Casey was supported by National Science Foundation (NSF) through the Mathematical Sciences Graduate Internship (MSGI) program. 
Anuradha Trivedi was supported in part by the U.S. Department of Energy, Office of Science, Office of Workforce Development for Teachers and Scientists (WDTS) under the Science Undergraduate Laboratory Internship (SULI) program, and in part by the Office of Advanced Scientific Computing Research, of the U.S. Department of Energy under Contract No. DE-AC02-05CH11231, through the grant ``Scalable Data-Computing Convergence and Scientific Knowledge Discovery", which also partially supported Talita Perciano.

This research used resources of the National Energy Research Scientific Computing Center (NERSC), a U.S. Department of Energy Office of Science User Facility located at Lawrence Berkeley National Laboratory, operated under Contract No. DE-AC02-05CH11231.

\small
\bibliographystyle{IEEEtranN}
\bibliography{references}
\end{document}